\documentclass[10pt,twocolumn,letterpaper]{article}

\usepackage[pagenumbers]{cvpr} 

\usepackage{graphicx}
\usepackage{amsmath}
\usepackage{amssymb}
\usepackage{booktabs}
\usepackage{balance}
\usepackage{lipsum}
\usepackage{caption}
\usepackage{algorithm}
\usepackage{algorithmic}
\usepackage{enumerate}
\usepackage{amsmath,amsthm,amssymb}
\usepackage{engord}
\usepackage{bm}
\usepackage{bbm}
\usepackage{amstext}
\usepackage{comment}
\usepackage{color}
\usepackage[x11names,table]{xcolor}
\usepackage{ctable}
\usepackage{booktabs}
\usepackage{multirow}
\usepackage{mathtools}
\usepackage{setspace}
\usepackage{pifont}
\usepackage{dsfont} 
\usepackage[inline]{enumitem}
\usepackage[toc,page]{appendix}
\usepackage[pagebackref,breaklinks,colorlinks]{hyperref}
\usepackage[capitalize]{cleveref}
\usepackage{hhline}

\usepackage[accsupp]{axessibility}

\definecolor{Gray}{gray}{0.9}
\definecolor{White}{gray}{1}
\definecolor{WhiteGray}{rgb}{0.9, 0.9, 0.9}
\definecolor{DGray}{gray}{0.8}
\definecolor{DDDDGray}{gray}{0.3}
\definecolor{citecolor}{HTML}{0071bc}

\definecolor{DeltaColor}{rgb}{0.039,0.73,0.71}
\definecolor{SigmaColor}{rgb}{0.98,0.45,0.0}
\definecolor{AlphaColor}{rgb}{0,0,0.8}
\definecolor{BetaColor}{rgb}{0.8,0,0.8}
\definecolor{GammaColor}{rgb}{0.514,0.34,0.224}
\definecolor{EpsilonColor}{rgb}{0.353,0.725,0.906}
\definecolor{GreenColor}{rgb}{0.137,0.573,0.565}
\definecolor{RedColor}{rgb}{0.949,0.275, 0.224}

\DeclareMathAlphabet\mathbfcal{OMS}{cmsy}{b}{n}

\newcommand{\colorRef}[1]{\textcolor{red}{#1}}
\crefname{figure}{\colorRef{Fig.}}{\colorRef{Figs.}}
\Crefname{figure}{\colorRef{Figure}}{\colorRef{Figures}}
\crefname{section}{\colorRef{Sec.}}{\colorRef{Secs.}}
\Crefname{section}{\colorRef{Section}}{\colorRef{Sections}}

\crefname{table}{\colorRef{Tab.}}{\colorRef{Tabs.}}
\Crefname{table}{\colorRef{Table}}{\colorRef{Tables}}
\Crefname{equation}{\colorRef{Eq.}}{\colorRef{Eqs.}}
\Crefname{equation}{\colorRef{Equation}}{\colorRef{Equation}}

\def\cvprPaperID{6805} 
\def\confName{CVPR}
\def\confYear{2023}

\newcommand\supp{\textbf{Appx}\xspace}
 
\newcommand\X{$\mathbf{X}$\xspace}
\newcommand\V{$\mathbf{V}$\xspace}
\newcommand\Pcloud{$\mathbfcal{PC}$\xspace}
\newcommand\Vcloud{$\mathbfcal{VC}$\xspace}
\newcommand\pt{$\mathbf{P}$\xspace}
\newcommand\pf{$\mathbf{F}$\xspace}
\newcommand\ptinball{$\mathbf{\bar{P}}$\xspace}
\newcommand\feat{$\mathbf{F}_{2d}$\xspace}
\newcommand\featprime{$\mathbf{F}^{\prime}_{2d}$\xspace}
\newcommand\featreshape{$\mathbf{F}_{3d}$\xspace}
\newcommand\meshgrid{$\mathbf{M}_{3d}$\xspace}
\newcommand\query{$\mathbf{Q}$\xspace}

\newcommand\I{$\mathbfcal{I}$\xspace}
\newcommand\world{$\mathcal{W}$\xspace}
\newcommand\method{POEM\xspace}
\newcommand\mv{\textit{m.v.}\xspace}
\newcommand\dexycb{DexYCB-MV\xspace}

\newcommand{\projectURL}{\href{https://github.com/lixiny/POEM}{\tt{github.com/lixiny/POEM}}}

\newcommand{\linenb}[1]{\noindent\mbox{\small\texttt{#1}}}
\newcommand{\cqheading}[1]{\vspace{1pt}\noindent\mbox{\textbf{#1}\;}}
\newcommand{\qheading}[1]{\vspace{5pt}\noindent\mbox{\textbf{#1}\;}}
\newcommand{\subqheading}[1]{\vspace{5pt}\noindent\mbox{\textit{\textbf{#1}}\;}}

\newcommand{\del}[1]{\xspace}
\newcommand{\qparagraph}[1]{\vspace{-1.0em}\paragraph*{#1}\mbox{}}
\newcommand{\alphapara}[1]{{\textcolor{red}{-#1}}}

\newcommand{\customfootnotetext}[2]{{
  \renewcommand{\thefootnote}{#1}
  \footnotetext[0]{#2}}}

\newlength\savewidth\newcommand\shline{\noalign{\global\savewidth\arrayrulewidth
  \global\arrayrulewidth 1pt}\hline\noalign{\global\arrayrulewidth\savewidth}}

\begin{document}

\title{
    \vspace{-0.5em}
    POEM: Reconstructing Hand in a Point Embedded Multi-view Stereo
    \vspace{-0.5em}
}

\author{
    {
        Lixin Yang\textsuperscript{1,2}\;
        Jian Xu\textsuperscript{3}\;
        Licheng Zhong\textsuperscript{1}\;
        Xinyu Zhan\textsuperscript{1}\;
        Zhicheng Wang\textsuperscript{3}\;
        Kejian Wu\textsuperscript{3}\;
        Cewu Lu\textsuperscript{1,2$\boldsymbol{\dagger}$}
    } \\
    { 
        \small
        {$^{1}$Shanghai Jiao Tong University}\quad\quad
        {$^{2}$Shanghai Qi Zhi Institute}\quad\quad
        {$^{3}$Nreal}
    }
    \\
    {
        \tt\small  \{{siriusyang}, {zlicheng}, {kelvin34501}, {lucewu}\}@{sjtu.edu.cn} 
    } \\
    {
        \tt\small  \{{jianxu}, {kejian}\}@{nreal.ai} \quad {chgggo}@{gmail.com}
    }\\
}
\maketitle
\customfootnotetext{$\dagger$}{
    Cewu Lu is the corresponding author, the member of Qing Yuan Research Institute and MoE Key Lab of Artificial Intelligence, AI Institute, Shanghai Jiao Tong University, China and Shanghai Qi Zhi institute.
}

\begin{abstract}
    Enable neural networks to capture 3D geometrical-aware features is essential in multi-view based vision tasks. 
    Previous methods usually encode the 3D information of multi-view stereo into the 2D features. 
    In contrast, we present a novel method, named \method, that directly operates on the 3D \textbf{PO}ints \textbf{E}mbedded in the \textbf{M}ulti-view stereo for reconstructing hand mesh in it.
    Point is a natural form of 3D information and an ideal medium for fusing features across views, as it has different projections on different views. Our method is thus in light of a simple yet effective idea, that a complex 3D hand mesh can be represented by a set of 3D points that 1) are embedded in the multi-view stereo, 2) carry features from the multi-view images, and 3) encircle the hand.  To leverage the power of points, we design two operations: point-based feature fusion and cross-set point attention mechanism. Evaluation on three challenging multi-view datasets shows that \method outperforms the state-of-the-art in hand mesh reconstruction. 
    Code and models are available for research at \projectURL.
\end{abstract}

\section{Introduction}
\label{sec:introduction}

Hand mesh reconstruction plays a central role in the field
of augmented and mixed reality, as it can not only deliver realistic experiences for the users in gaming but also support applications involving teleoperation, communication, education, and fitness outside of gaming. Many significant efforts have been made for the monocular 3D hand mesh reconstruction \cite{boukhayma20193d,Park2022HandOccNet, ge20193d,chen2021camera,moon2020i2l,chen2021i2uv}.
However, it still struggles to produce applicable results, mainly for these three reasons. 
\begin{enumerate*}[label={\textbf{(\arabic*)}}]
    \item \textit{Depth ambiguity.} Recovery of the absolute position in a monocular camera system is an ill-posed problem. Hence, previous methods \cite{ge20193d,zhou2020monocular,moon2020i2l} only recovered the hand vertices relative to the wrist  (\ie root-relative). 
    \item \textit{Unknown perspectives.} The shape of the hand’s 2D projection is highly dependent on the camera’s perspective model (\ie camera intrinsic matrix). However, the monocular-based methods usually suggest a weak perspective projection  \cite{boukhayma20193d, lin2021metro}, which is not accurate enough to recover the hand’s 3D structure. 
    \item \textit{Occlusion.} The occlusion between the hand and its interacting objects also challenges the accuracy of the reconstruction  \cite{Park2022HandOccNet}. 
\end{enumerate*}
These issues limit monocular-based methods from practical application, in which the absolute and accurate position of the hand surface is required for interacting with our surroundings.

\begin{figure}[!t]
    \begin{center}
      \includegraphics[width=0.80\linewidth]{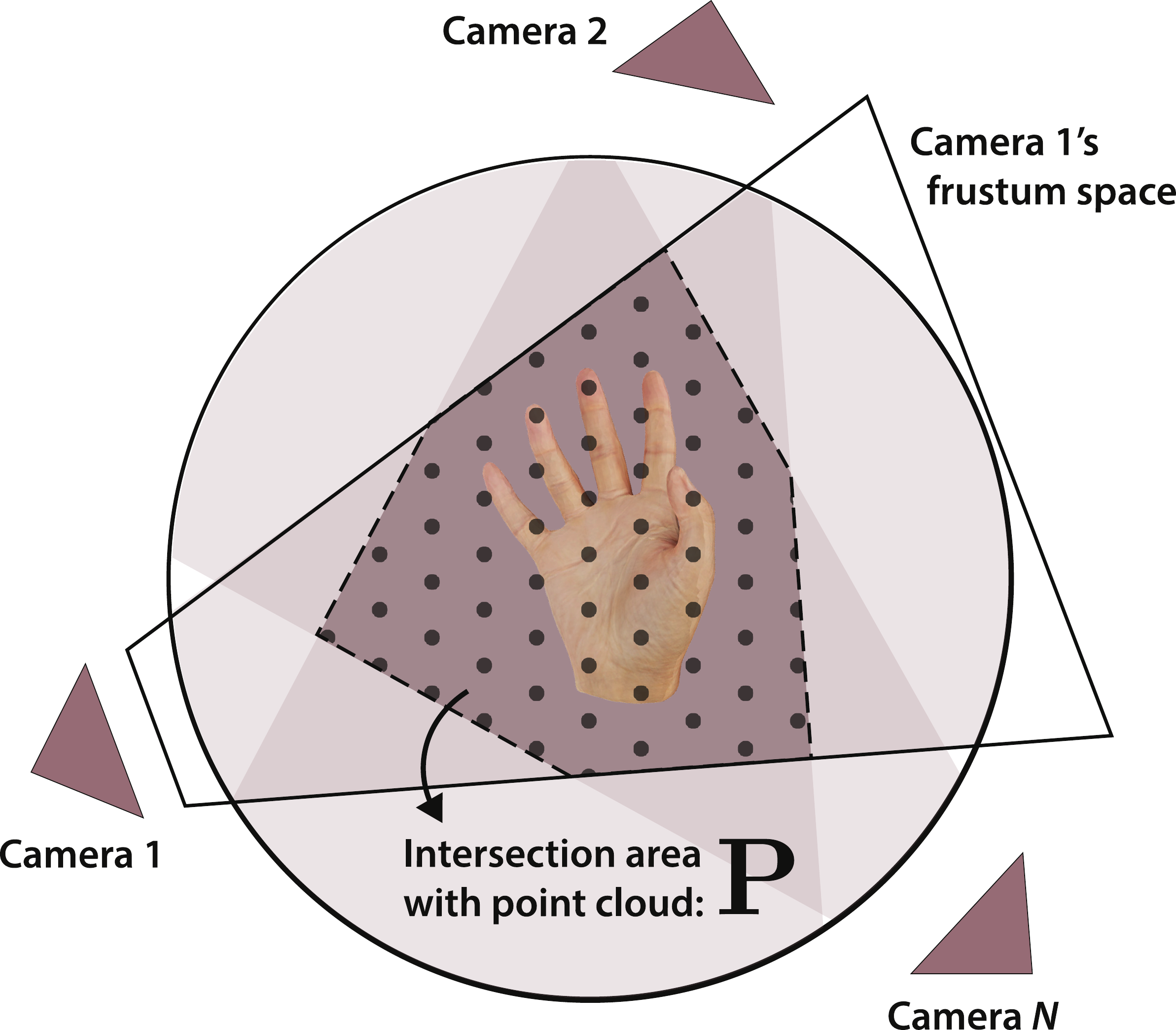}
    \end{center}\vspace{-15pt}
      \caption{\textbf{Intersection area of \textit{N} cameras' frustum spaces.}
      The gray dots represent the point cloud \pt aggregated from \textit{N} frustums. Our method: \method, standing for the point embedded multi-view stereo, focuses on the dark area scatted with gray dots.}
      \vspace{-12pt}
    \label{fig:multi_cams_views}
\end{figure}

Our paper is thus focusing on reconstructing hands from multi-view images.  
Motivation comes from two aspects. 
First, the issues mentioned above can be alleviated by leveraging the geometrical consistency among multi-view images.
Second, the prospered multi-view hand-object tracking setups \cite{brahmbhatt2020contactpose,Chao2021DexYCB,Yang2022OakInk, zimmermann2019freihand} and VR headsets bring us an urgent demand and direct application of multi-view hand reconstruction in real-time. 
A common practice of multi-view 3D pose estimation follows a two-stage design.  It first estimates the 2D key points of the skeleton in each view and then back-project them to 3D space through several 2D-to-3D lifting methods, \eg algebraic triangulation \cite{He2020EpipolarT, Iskakov2019LearnableT, Remelli2020LightweightM3}, Pictorial Structures Model (PSM) \cite{Pavlakos2017HarvestingMV, Qiu2019CrossVF}, 3D CNN \cite{Tu2020VoxelPose, Iskakov2019LearnableT}, plane sweep \cite{Lin2021mvp}, etc.
However, these two-stage methods are not capable of reconstructing an animatable hand mesh that contains both skeleton and surface.
It was not until recently that a one-stage multi-view mesh regression model was proposed \cite{Wang2021MVP}. 

How to effectively fuse the features from different images is a key component in the multi-view setting. 
Accordingly, previous methods can be categorized into three types. 
\begin{enumerate*}[label={\textbf{(\arabic*)}}]
    \item \textbf{Fusing in 2D.} The features are directly fused in the 2D domain using explicit epipolar transform \cite{Qiu2019CrossVF,He2020EpipolarT} or implicit representations that encode the camera transformation  (\ie camera intrinsic and extrinsic matrix) into 2D features, \eg feature transform layer (FTL) \cite{Remelli2020LightweightM3, Han2022UmeTrack} and 3D position embedding (RayConv) \cite{Wang2021MVP};
    \item \textbf{Fusing in 3D.} The features are fused in a 3D voxel space via PSM \cite{Qiu2019CrossVF, Pavlakos2017HarvestingMV} or 3D CNNs \cite{Iskakov2019LearnableT, Tu2020VoxelPose};
    \item \textbf{Fusing via 3D-2D projection.} The features are fused by first projecting the 3D keypoints’ initial guess into each 2D plane and then fusing multi-view features near those 2D locations \cite{Wang2021MVP};
\end{enumerate*}

The fusion mode in type 1 is considered as \textit{holistic}, since it indiscriminately fuses all the features from different views. Consequentially, it ignores the structure of the underlying hand model that we are interested in. 
On the contrary, the fusion mode in type 3 is considered as \textit{local}. 
However, only the features around the 2D keypoints are hard to capture the consistent geometrical features from a global view. Besides, the 3D keypoints initial guess may not be accurate enough, resulting in the fusion being unstable.
The fusion mode in type 2 is not in our consideration as it tends to be computationally expensive and suffers from 
quantization error.

Based on the above discussion, we aim to seek a feature representation and a fusion mode between type 1 and type 3 for both holistically and locally fusing the features in multi-views, and to explore a framework for robust and accurate hand mesh reconstruction. Our method is called \method, standing for \textit{\textbf{PO}int \textbf{E}mbedded \textbf{M}ulti-view Stereo}. We draw inspiration from the Basis Point Set (BPS) \cite{prokudin2019bps}, which bases on a simple yet effective idea that a complex 3D shape can be represented by a fixed set of points (BPS) that wraps the shape in it. 
If we consider the intersection of different cameras' frustum spaces as a point cloud, and the hand's vertices as another point cloud, then the intersected space is the basis point set for hand vertices (see \cref{fig:multi_cams_views}).
Once we assign the multi-view image features to the point cloud in the intersected space, fusing image features across different views becomes fusing the point features from different camera frustums. 
The advantages of this representation are two-fold:
\begin{enumerate*}[label={\textbf{(\roman*)}}]
    \item The hand is wrapped in a dense point cloud (set) that carries dense image features collected from different views, which is more holistic and robust than the local fusion mode in type 3.
    \item For each vertex on the hand surface, it interacts with basis points in its local neighborhood (\ie $k$ nearest neighbor), which is more selective than the holistic fusion mode in type 1.
\end{enumerate*}

\cref{fig:architecture} shows our model's architecture.
\method consists of two stages. 
In the first stage (\cref{sec:keypoints_estimation}), \method takes the multi-view images as input and predicts the 2D keypoints of the hand skeleton in each view.
Then, the 3D keypoints are recovered by an algebraic triangulation module. 
In the second stage, \method fuses the features from different views in a space embedded by points
and predicts the hand mesh in this space  (\cref{sec:mesh_estimation}). 
The point feature on hand vertices will iteratively interact with the features of the embedded points through a cross-set attention mechanism, and the updated vertex features are further used by \method to predict the vertex's refined position (\cref{sec:cross_set_point_transformer}). 

We conduct extensive experiments on three multi-view datasets for hand mesh reconstruction under the object's occlusion, 
namely HO3D \cite{hampali2021ho3dv3}, DexYCB \cite{Chao2021DexYCB}, and OakInk \cite{Yang2022OakInk}. With the proposed fusion mode and attention mechanism, \method achieves state-of-the-art on all three datasets.

Our contributions are in three-fold:
\begin{itemize}[leftmargin=12.pt]
    \setlength\itemsep{-4 pt}
    \vspace{-6pt}
    \item We investigate the multi-view pose and shape reconstruction problem from a new perspective, that is, the interaction between a target point set (\ie mesh vertices) and a basis point set (\ie point cloud in the camera  frustum spaces). 
    \item According to that, we propose an end-to-end learning framework: \method for reconstructing hand mesh from multi-view images through a point embedded multi-view stereo. To encourage interaction between two point sets, \method introduces two new operations: a point-based feature fusion strategy and a cross-set point attention. 
    \item We conduct extensive experiments to demonstrate the efficacy of the architecture in \method. As a regression model targeting mesh reconstruction, \method achieves significant improvement compared to the previous state-of-the-art.    
\end{itemize}

\section{Related Work}
\vspace{-6pt}
\qheading{Multi-view Feature Processing.}
Representing the observations from different camera systems in a unified way while fusing multi-view features accordingly is a common problem in multi-view stereo (MVS) reconstruction and pose estimation. This literature review focuses on addressing this key challenge.
From this perspective, previous methods - in which the camera transformation (extrinsic) is typically encoded differently - can be seen as different types of Position Embedding (PE).
For example, the method based on epipolar transform \cite{He2020EpipolarT} can be classified as a line-formed PE, as pixels in one camera are encoded as epipolar lines in others. 
Additionally, there are point-formed position embeddings such as FTL in \cite{Remelli2020LightweightM3, Han2022UmeTrack}, RayConv in \cite{Wang2021MVP}, and 3D Position Encoder in PETR \cite{Liu2022PETR}, which apply camera extrinsic directly to point-shaped features (FTL) or add camera ray vectors channel-wise to the features (RayConv, PETR).
These two point-formed PEs use point-formed features solely in 2D format because 2D convolution or image-based self-attention cannot capture 3D structure. 
Therefore they are considered implicit. 
In contrast, 3D CNN and point cloud network can preserve 3D structure. SurfaceNet \cite{ji2017surfacenet} and LSM \cite{kar2017LSM} associate features from different views by forming a cost volume and rely on 3D CNNs to perform voxel-wise reconstruction. To address the drawback of the final volumetric output, MVSNet \cite{yao2018mvsnet} predicts the depth-map instead of voxels. These methods are classified as voxel-formed PEs.
Finally, the explicit point-formed PE directly uses a set of 3D points in the MVS scene.
For instance, PointMVS \cite{chen2019PointMVSNet} unprojects the predicted depth-map to a point cloud and aggregates features from different views using project-and-fetch. Our method also belongs to this type.
In our task, preserving the topology of the hand vertex points is crucial, but this can be challenging with PointMVS, which indiscriminately treats points. Instead, our approach, \method, represents the common-view scene as an unstructured point cloud for feature aggregation and employs a structure-aware vertex query to initialize and update hand vertices. \method's structured vertices interact with unstructured frustum points through cross-attention, which effectively removes object occlusion and accurately reconstructs the hand mesh.

\vspace{-0.5em}\qheading{Monocular Hand Reconstruction.}
Monocular hand reconstruction has been a long-studied topic. A series of works \cite{boukhayma20193d,hasson2019learning, zhou2020monocular, Kong2022IdentityAwareHM} were built upon the deformable hand mesh with a differentiable skinning function, \textit{e}.\textit{g}.\  MANO \cite{romero2017embodied}. 
However, the difficulty of regressing the non-Euclidean rotations hinders the performance of these methods.
There have been emerging works exploring the direct reconstruction of the hand surface. As vertices naturally lie in 3D Euclidean space, 
\cite{kolotouros2019convolutional,ge20193d,kulon2020weakly,chen2021camera} leveraged the mesh structure of MANO with the graph-based convolution networks (GCN). Besides, the voxels \cite{moon2020i2l}, UV positional maps \cite{chen2021i2uv}, and signed distance function \cite{karunratanakul2020grasping,chen2022alignsdf} were also competent choices. Recently, Transformers \cite{lin2021metro,lin2021graphormer} have been deployed to fuse the features on the hand surface by self-attention mechanism. 
In this work, we follow the path of direct mesh reconstruction with Transformer and propose to model hand mesh as a \textbf{point set} in the multi-view stereo. 

\vspace{-0.5em}\qheading{Point Cloud Processing.}
Qi \etal \cite{qi2017pointnet} proposed PointNet, the first deep model utilizing the permutation-invariant structure of point cloud data. A subsequent work,  PointNet++, was proposed in \cite{qi2017pointnet++} thereafter. PointNet++ introduced ball query and hierarchical grouping, enabling the model to reason form local structures of point clouds. There were a number of successive works \cite{li2018pointcnn,wu2019pointconv,thomas2019kpconv} trying to define local convolution operators on point clouds to extract local information. 
The recent application of transformers on point cloud data has been proven a success. Zhao \textit{et al}.\  proposed Point Transformer \cite{Zhao2021PointT}, which adopts a vector attention mechanism to perform attention in the local neighborhood. Guo \textit{et al}.\  proposed the Point Cloud Transformer \cite{guo2021pct}, which used an analogy of the Laplacian matrix on point clouds to fuse long-range relationships in the point cloud.
Our method follows the design in Point Transformer, but selectively fuses the points from different cameras to a hand vertex through a \textbf{cross-set} vector attention.

\vspace{-0.5em}
\section{Method}
\begin{figure*}[ht]
    \begin{center}
      \includegraphics[width=0.95\linewidth]{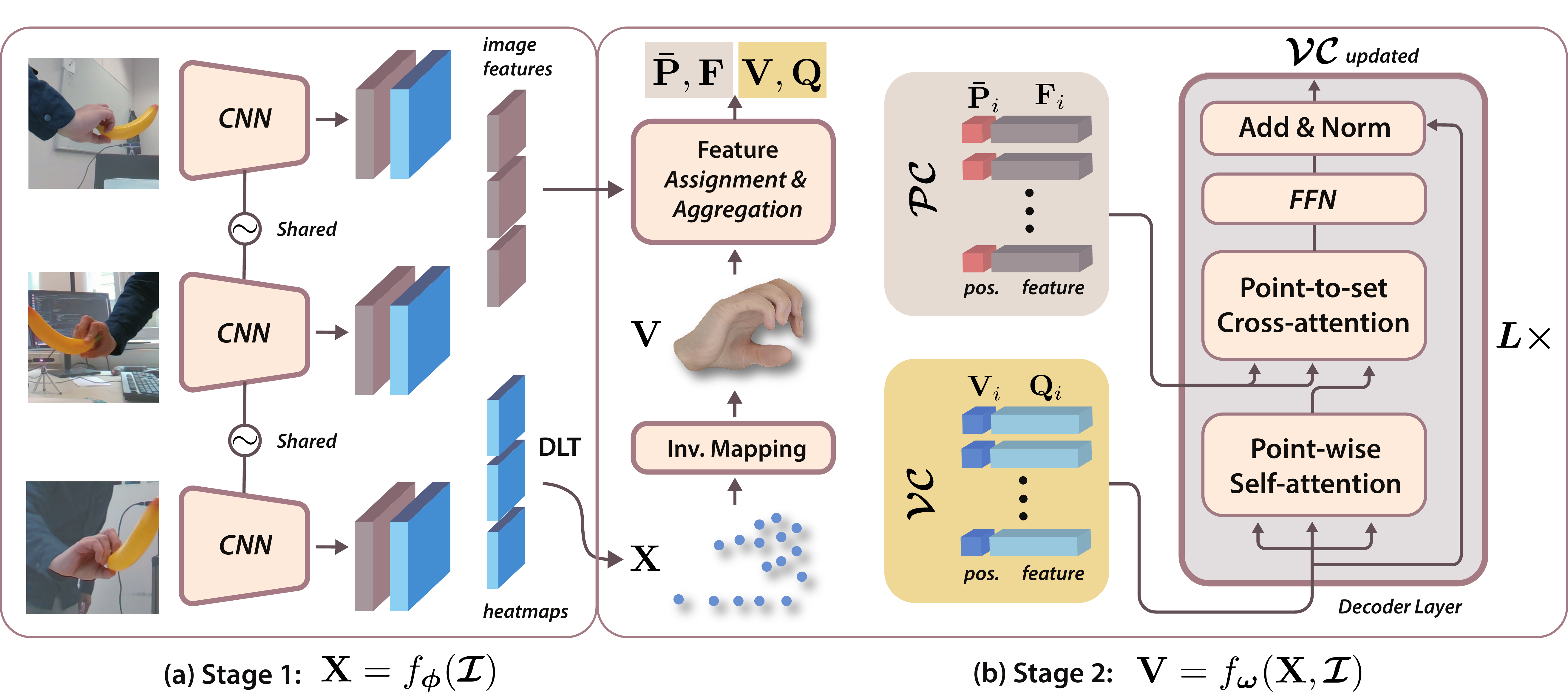}
    \end{center}\vspace{-15pt}
      \caption{\method's architecture contains two stages: (a) hand's keypoints prediction conditioned on multi-view images, and (b) hand's vertices prediction conditioned on the predicted keypoints and multi-view images features.}\vspace{-1em}
    \label{fig:architecture}
\end{figure*}
\subsection{Formulation}
\label{sec:formulation}
The general purpose of this paper is to model the joint distribution of hand skeleton and surface under multi-view observations.
Given \textit{N} cameras with different positions and orientations, we first define the 3D coordinates of the hand's skeleton and surface inside a shared \textit{3D world space}: \world.
Accordingly, their coordinates inside each camera frame can be retrieved by applying extrinsic transformation. 
Let $\mathbf{X} \in \mathbb{R}^{21\times3}$ denotes the 3D keypoints on the hand skeleton, and let $\mathbf{V} \in \mathbb{R}^{778\times3}$ denotes the 3D vertices on the hand surface mesh,
given a image set $\mathbfcal{I} = \{\mathbf{I}_v\}^{N}_{v=1}$ from total \textit{N} views, the proposed model predicts a distribution
$P(\mathbf{X}, \mathbf{V} | \mathbfcal{I})$.
However, directly model this joint distribution is hard and suffers from inferior performance (see Exp. \cref{para:poem_two_stage}\alphapara{B}). 
Hence, based on the chain rule, we decompose the joint distribution as follows:
\begin{equation}
    \setlength{\abovedisplayskip}{5pt}
    \setlength{\belowdisplayskip}{5pt}
    P(\mathbf{X}, \mathbf{V} | \mathbfcal{I}) = P_{\bm{\omega}}(\mathbf{V} | \mathbf{X}, \mathbfcal{I}) P_{\bm{\phi}}(\mathbf{X} | \mathbfcal{I}).
    \label{eq:formulation}
\end{equation}
This equation expresses \method's architecture (shown in \cref{fig:architecture}), which consists of two stages.
The training process is to fit the learnable parameters $\bm{\phi}, \bm{\omega}$ on the training data, \ie
for each input $\mathbfcal{I}$, the model maximizes its probability at the ground-truth $\mathbf{\widehat{X}}, \mathbf{\widehat{V}}$.
We represent these two stages: $P_{\bm{\phi}}$ and  $P_{\bm{\omega}}$ as two neural networks: $f_{\bm{\phi}}(\mathbfcal{I})$ and $f_{\bm{\omega}}(\mathbf{X}, \mathbfcal{I})$. The fitting objective are set to the standard $\mathit{l}$1 loss:
\begin{equation}
    \setlength{\abovedisplayskip}{5pt}
    \setlength{\belowdisplayskip}{5pt}
    \mathcal{L} = \big| \mathbf{\widehat{V}} - f_{\bm{\omega}}(\mathbf{X}, \mathbfcal{I}) \big| + \big| \mathbf{\widehat{X}} - f_{\bm{\phi}}(\mathbfcal{I})\big|.
    \label{eq:overall_objectives} 
\end{equation}

Having outlined the problem decomposition and learning objectives, we now move on to discuss details of the network design.

\subsection{Keypoints from M.v. Images}
\label{sec:keypoints_estimation}
The first stage of \method is to predict the 3D keypoints \X conditioned on the multi-view (\mv) images \I.
To achieve this, 
we first estimate the $\mathbf{X}$'s 2D location in each view $\mathbf{I}_i$ and
then lift them to the 3D space through algebraic triangulation. 
To increase 2D keypoints' robustness against occlusion, we leverage 2D likelihood heatmap \cite{Wei2016CPM} and \textit{soft-argmax} operation \cite{sun2018integral} for $\mathbf{X}$'s 2D location.
Inside the first stage, we suppose that the 2D keypoints from each view's prediction have independent contribution to $\mathbf{X}$. 
Hence we can apply the direct linear transformation (DLT, \cite[p.312]{hartley2003MVGeometry})  for fast triangulation. Formally, \method's first stage $f_{\bm{\phi}}(\mathbfcal{I})$ can be expressed as:
\begin{equation}
    \setlength{\abovedisplayskip}{5pt}
    \setlength{\belowdisplayskip}{5pt}
    \begin{aligned}
    f_{\bm{\phi}}: \ &\mathbf{X} = \text{DLT}(\mathbf{h}_{1\sim N}, \mathbf{K}_{1\sim N}, \mathbf{T}_{1 \sim N} ), \\ &\text{where} \ \mathbf{h}_i = \textit{soft-argmax}\ (\mathcal{F}_{\bm{\phi}}(\mathbf{I}_i)),
    \label{eq:keypoints_estimation} 
    \end{aligned} 
\end{equation}
in which $\mathcal{F}_{\bm{\phi}}$ is a backbone network with weights $\bm{\phi}$, $\mathbf{K}_{i}$, $\mathbf{T}_{i}$ is the camera intrinsic and extrinsic matrix of $i$-th camera, respectively.

\subsection{Vertices from Keypoints and M.v. Images}
\label{sec:mesh_estimation}
In the previous stage, we built a model that can robustly retrieve the hand's keypoints \X from multi-view images. 
However, only the keypoints \X is deficient for our task in two aspects: 
\begin{enumerate*}[label={\textbf{(\arabic*)}}]
    \item keypoints are not on the surface, which can hardly reflect the shape of hand; 
    \item the keypoints are retrieved independently from the backbone model, lacking the information fused from different views;
\end{enumerate*}
Based on (2), the keypoints may not be accurate enough.  

Given such deficiencies, inside the second stage, we want the latent embeddings of hand vertices \V can fully interact with 
the features from all views before reaching its outputs: \V.
We start by reviewing the problem setting that each image observes a frustum space in front of a camera. 
As shown in \cref{fig:multi_cams_views}, the common views among all \textit{N} cameras reflect the \textit{intersection} area of \textit{N} frustum spaces. The hand that we are
interested in  lies in it.  

To enable the network to operate on the frustum space, we discretize it into a 3D meshgrid. 
Each cell in the meshgrid store a 3D coordinate of a point. With the camera extrinsic, 
we can transform all the points from total \textit{N} camera frustums into the shared world space \world (details in \cref{sec:cam_frustum_space}).
Therefore, the hand is surrounded by a point cloud \pt aggregated from \textit{N} frustums. 
Now let us revisit the form of \V. If we consider the 778 vertices of \V as another point cloud in \world, 
the goal of the second stage becomes encouraging the interaction between two point clouds: 
One from \textit{N} camera frustums, the other from the surface of hand.

Representing the camera frustum and hand vertices in point cloud has two advantages. 
\begin{enumerate*}[label={\textbf{(\arabic*)}}]
    \item Point cloud is invariant to the permutation of points. It only subject to the relative distance between points. 
    Hence, points from different camera frustums can be easily fused to the points in \V if they are spatially close;
    \item Point cloud is a set of 3D points with its coordinates as a natural positional encoding, which is quite suitable for 
    the mechanism of self and cross-attention;
\end{enumerate*}
Given these advantages, we design a cross-set point Transformer to fuse features in point cloud (\cref{sec:cross_set_point_transformer}).

One last issue is how to embed features into the point clouds \pt and \V, since currently they are just a set of 3D coordinates.
We investigate two ways in the \cref{sec:feats_to_points_assignment}. 
We start by introducing a \textit{Position Embedded Aggregation} and followed by a more powerful \textit{Projective Aggregation}.

\begin{figure*}[ht]
    \begin{center}
        \begin{minipage}[ht]{0.39\textwidth}
        \centering
        \resizebox{0.95\linewidth}{!}{
            \includegraphics[width=1.0\linewidth]{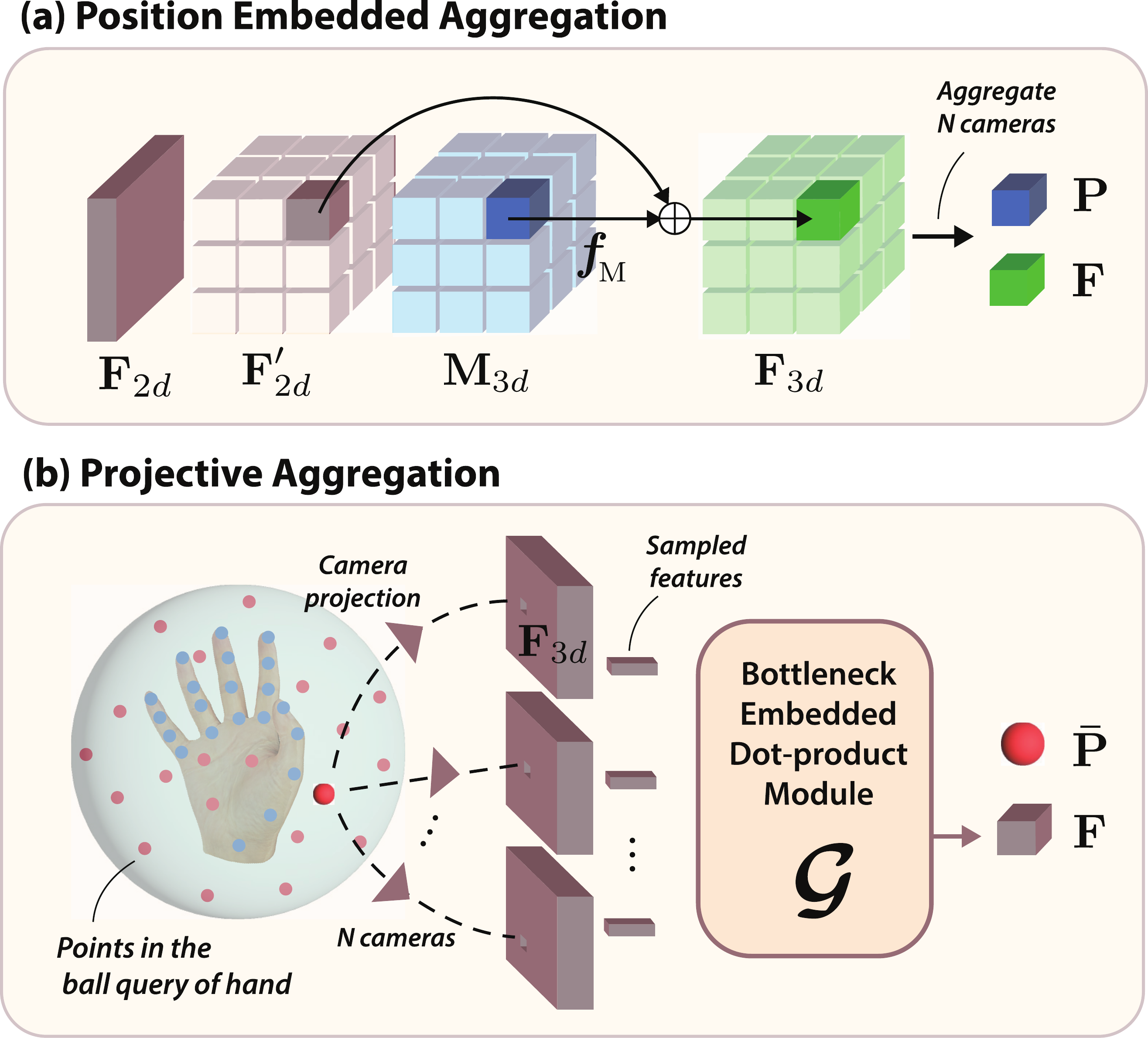}
            }\vspace{-4pt}
            \caption{Illustration of the (a) position embedded aggregation and (b) projective aggregation for embedding image features into point cloud. }
            \label{fig:feature_aggregation}
        \end{minipage}
        \quad
        \begin{minipage}[htp]{0.28\textwidth}
        \centering
            \resizebox{\linewidth}{!}
            {
                \includegraphics[width=1.0\linewidth]{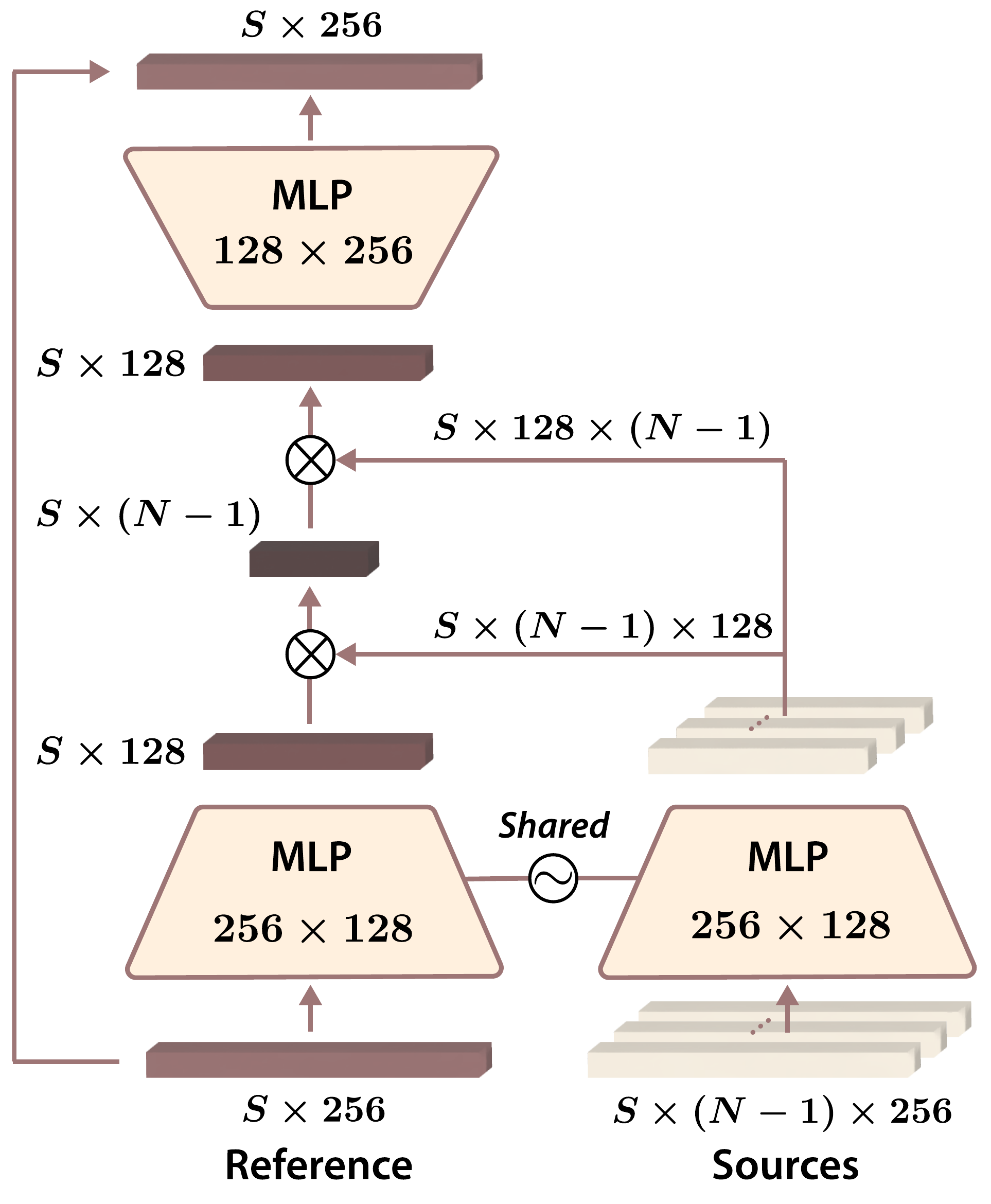}
            }\vspace{-4pt}
            \caption{Architecture of $\mathcal{G}$: bottleneck embedded dot-product module for fusing features from different views.}
            \label{fig:bottleneck_embed_dotproduct}
        \end{minipage}
        \quad
        \begin{minipage}[htp]{0.28\textwidth}
        \centering
            \vspace{5pt}
            \resizebox{1.0\linewidth}{!}
            {   
                \includegraphics[width=1.0\linewidth]{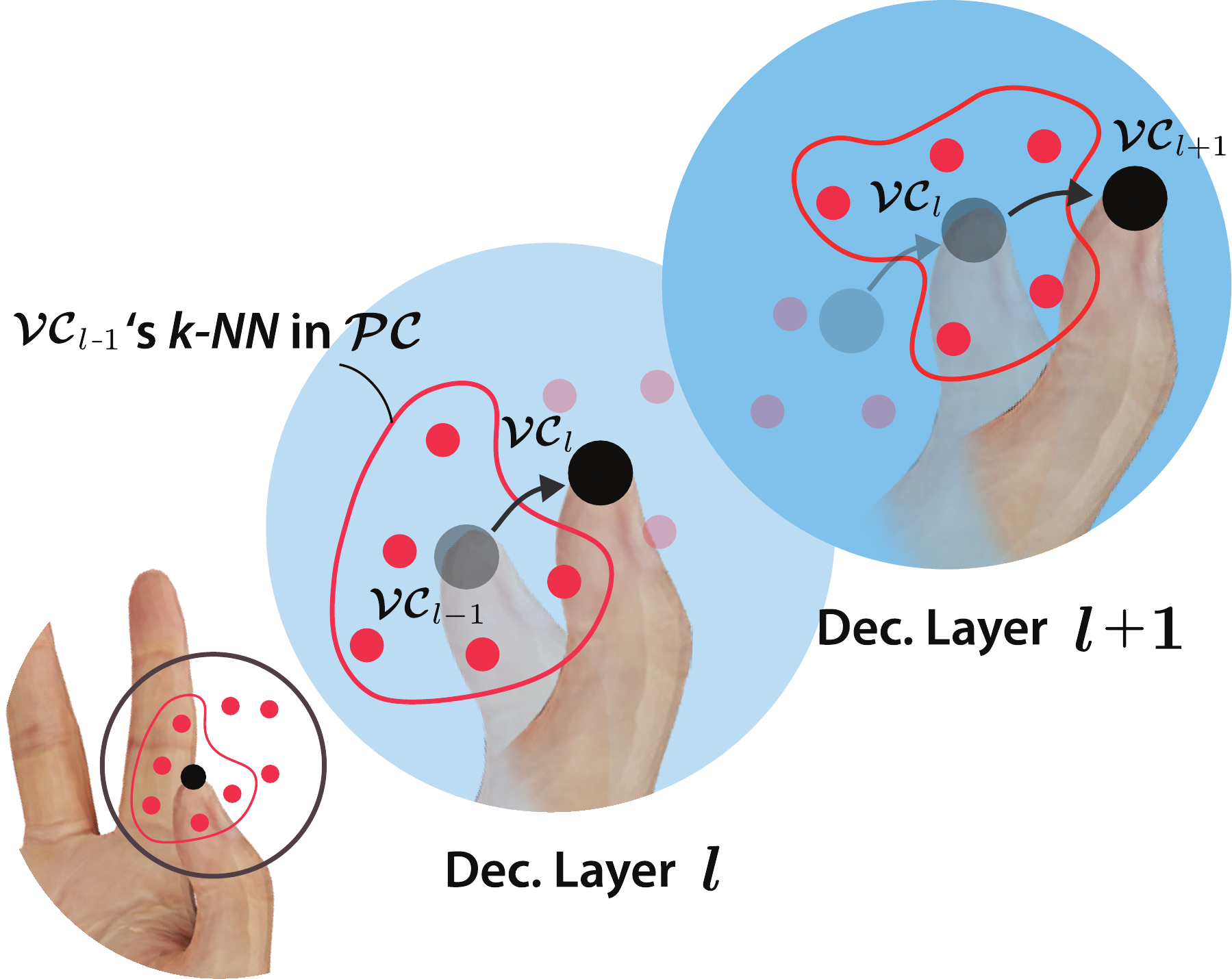}
            }\vspace{2pt}
            \caption{Vector attention: in each decoder layer, the \Vcloud's position will be updated by the attention between \Vcloud and  \Vcloud's $k$ nearest neighbors in \Pcloud.
            Accordingly, \Vcloud's $k$ nearest neighbors will be updated based on refined position of \Vcloud.}
            \label{fig:transformer_decoder_knn}
        \end{minipage}\vspace{-2em}
    \end{center}
\end{figure*}

\vspace{-1.0em}
\subsubsection{Embedding Points to Camera Frustum Space}
\vspace{-0.3em}
\label{sec:cam_frustum_space}

To associate each camera frustum into a shared world space \world,
we propose to first embed discrete points along the camera ray direction in the frustum and then transform the embedded points to \world. 
A camera ray can be represented by a 3D vector $\bm{r} = (u, v, d)$, where $u, v$ are the pixel coordinates in the image plane, and $d$ is a depth value.
Accordingly, all the camera ray vectors constitute a meshgrid of size $W \times H \times D$, where $W, H$ are the width and height of the image, and $D$ is the number of discrete depth values.
Finally, given the camera intrinsic matrix $\mathbf{K}$ and extrinsic matrix $\mathbf{T}$ (from camera to \world), 
we can transfer the points in each camera's meshgrid to \world through 
firstly back projecting the ray vector $\bm{r}$ into a 3D point in camera space using $\mathbf{K}$, and then transform the point to \world using $\mathbf{T}$.
After aggregating points from all \textit{N} cameras, we obtain the final point cloud \pt of size $M=N \times W \times H \times D$, which represents a wide range of points surrounding the hand \V. 

\smallskip
However, only points that are spatially close to \V are relevant. Ideally, we aim for the number of relevant points ($S$) to be greater than the number of vertices of hand ($S > 778$), while still being considerably smaller than $M$ ($S\ll M$). To achieve this, we need to determine the initial position of the surface vertex, \V, based on the predicted skeleton joints, \X. While MANO's formulation provides a pretrained mapping from \V to \X, mapping \X to \V is not straightforward. A basic approach for this inverse mapping is to learn a neural network {\small $f_m: \mathbf{X} \mapsto \mathbf{V}$}. However, we found that the initial {\small $\mathbf{V} = f_m(\mathbf{X})$} may deviate significantly from $\mathbf{X}$, resulting in unstable training. Therefore, instead of learning the absolute position of $\mathbf{V}$, we choose to learn $\mathbf{V}$ as an offset of the wrist (root) joint $\mathbf{X}_w$, as follows: {\small $\mathbf{V} = \mathbf{X}_w + f_m(\mathbf{X})$}, where $\mathbf{X}_w$ is the wrist joint in the world system. Once \V is obtained, we can retrieve its relevant points from \pt using the ball query \cite{qi2017pointnet++} operation. These relevant points lie within a certain radius of \V and are a subset of \pt. We refer to them as \ptinball.
\del{In the next section, we will introduce two variants for embedding image features into \ptinball.}

\vspace{-1.0em}
\subsubsection{Embeding Features to Points}
\vspace{-0.5em}
\label{sec:feats_to_points_assignment}
In this section, we discuss the approaches for embedding image features to camera frustum points (\pt \& \ptinball, \cref{para:feats_to_p_assignment}\alphapara{A}) and hand surface points (\V, \cref{para:feats_to_v_assignment}\alphapara{B}), respectively. 

\vspace{0.2em}
\qparagraph{A. Embedding Features to Camera Frustum Points (\pt)}\label{para:feats_to_p_assignment}
\vspace{-1.2em}

\subqheading{Position Embedded Aggregation.}\quad
Recall in \cref{sec:cam_frustum_space}, we use a meshgrid of 3D points to represent the camera ray vectors inside the camera frustum.
This meshgrid \meshgrid is a tensor of shape: $(W, H, D, 3)$. 
Meantime, the image feature \feat extracted from the backbone model is a 
tensor with different shape: $(C, H, W)$, where $C$ is the numbers of channel.  
A basic design to assign the features in \feat to the points in \meshgrid is 
through embedding the \meshgrid's positions into \feat, as shown in  \cref{fig:feature_aggregation}\alphapara{(a)}.
Specifically, we first reshape the \feat to \featprime: $(W, H, D, \frac{C}{D})$, 
where the last dimension (of size $\frac{C}{D}$) stores \feat's 3D position-aware components.  
Then, we add the \meshgrid's positional encoding and \featprime up for the \textit{position embedded} feature: \featreshape:
\begin{equation}
    \setlength{\abovedisplayskip}{5pt}
    \setlength{\belowdisplayskip}{5pt}
    \mathbf{F}_{3d} = f_{\mathrm{M}}(\mathbf{M}_{3d}) + \mathbf{F}^{\prime}_{2d},
    \label{eq:position_embedded_aggregation} 
\end{equation}
where $f_{\mathrm{M}}$ is a series of sinusoids and MLP functions to obtain \meshgrid' s positional encoding. 
The \featreshape has an identical spatial arrangement to \meshgrid{}. Each cell in \featreshape stores a feature corresponding to the point at the same position in \meshgrid{}. Therefore, the underlying idea for embedding image features into frustum points involves associating each point in \meshgrid{} with its corresponding feature in \featreshape{}.  Subsequently, following the application of ball query, \ptinball gathers points with different image features from different viewpoints, leading to effective feature aggregation.
\del{After performing the ball query operation, \ptinball takes in points with different image features from different views. As a result, the aggregation is achieved.}
This approach is akin to previous methods such as PETR \cite{Liu2022PETR} (referred to as "3D Positional Encoder") and MVP \cite{Wang2021MVP} (referred to as "Ray Convolution"), which also rely on similar feature embedding strategy.

Although the above design seems reasonable, it suffers from two drawbacks. 
Firstly, we select \ptinball instead of \pt, which results in only a small portion of feature cells in \featreshape{} being assigned to the point cloud \ptinball. Secondly, since each point in \ptinball only carries features from one camera frustum, the features in \ptinball lack consistency across different views

\del{First, since we choose \ptinball instead of \pt, only a small portion of feature cells in \featreshape are assigned to the point cloud \ptinball;
Second, since each point in \ptinball only carry feature from one camera frustum, the features in \ptinball lack consistency across different views;}

\subqheading{Projective Aggregation.}\quad
To overcome the aforementioned drawbacks and facilitate more effective feature aggregation, 
we propose fusing  features of points in \ptinball from different views. 
As shown in \cref{fig:feature_aggregation}\alphapara{(b)}, \ the key concept is to collect the features sampled at the 2D projection of a specific point in \ptinball across $N$ views, and combine the $N$ independently sampled features into a single geometry-aware feature for that point. 
We refer to this operation as \textit{projective aggregation}.
To sample feature at projected location  in \featreshape, we use bilinear interpolation. 
To fuse the features sampled from different camera views, we modify an original non-local network  \cite{Wang2018NonlocalNN} with bottleneck embedded dot-product design. As shown in \cref{fig:bottleneck_embed_dotproduct},
we choose the feature of one camera as reference, and the remain $N-1$ as sources. 
In the beginning, all the $N$ sampled features are downsampled by an embedding projection (a MLP function).
Then the features of sources and reference are completely mixed through dot-product, and up-sampled back by another embedding projection.
At last, the mixed features are added to the reference for the final fused feature. Formally, the projective aggregation can be expressed as:
\begin{equation}
    \setlength{\abovedisplayskip}{5pt}
    \setlength{\belowdisplayskip}{5pt}
    \mathbf{F} = \mathcal{G}(\mathbf{f}_1,...,\mathbf{f}_N), \;\text{where} \; \mathbf{f}_i = \mathbf{F}_{3d} \big( \pi(\bar{\mathbf{P}}) \big)_i
    \label{eq:projective_aggregation}
\end{equation}
where $\mathbf{F}_{3d} (\cdot)_i$ is bilinear interpolation on the $i$-th $\mathbf{F}_{3d}$, $\pi$ is the camera projection, and $\mathcal{G}$ is the bottleneck embedded dot-product module.

Compared to features of position embedded aggregation, the feature produced by projective aggregation is geometrical-aware. 
For example, two points from different camera frustums will have similar fused features if they are spatially close. 
On the other hand, two points that are close in one image plane will still have the fused features of significant difference, if they are spatially far away.

\qparagraph{B.\; Embedding Features to Hand Surface Points (\V)}\label{para:feats_to_v_assignment}\vspace{0.3 em}

{\noindent}Similar to the features of \ptinball, we gather the \V's multi-view image feature by projective aggregation.
In addition, since we are using \V as the token in the Transformer, 
providing positional encoding for \V is crucial.
Following the practices of several vision-based Transformers, such as Keypoint Transformer \cite{Hampali2022KeypointTS}, Mesh Transformer \cite{lin2021metro}, and MVP \cite{Wang2021MVP}, we utilize a set of learnable parameters: $\mathbf{S} \in \mathbb{R}^{N_V \times Z}$, as the position encoding for \V, where number of vertex: $N_V = 778$, and dimension of the encoding: $Z=256$. 
We explore three different strategies for constructing $\mathbf{S}$.
\vspace{0.1em}
\begin{enumerate}[label={\textbf{(\arabic*)}}, leftmargin=18pt]
    \setlength\itemsep{-2 pt}
    \item \textbf{J-emb} ($\mathbf{S}^{J}$). Similar to Keypoint Transformer \cite{Hampali2022KeypointTS}, the $\mathbf{S}^{J}$ is initialized as the joint/vertex-level learnable embedding vectors: $\mathbf{S}^{J} \in \mathbb{R}^{N_V \times Z}$.
    \item \textbf{G-emb} ($\mathbf{S}^{G}$). Similar to the Mesh Transformer \cite{lin2021metro}, the $\mathbf{S}^{G}$ is initialized as the concatenation ($\oplus$) of an input-dependent global image feature: $\mathbf{G} \in \mathbb{R}^{Z}$ and a constant vertex-specified position: $\mathbf{V}_i^{\dagger}$ ($i$ for $i$-th vertex) extracted from a zero-posed and mean-shape MANO hand template ($\dagger$). Therefore, $\mathbf{S}_i^{{G}} = \mathbf{G} \oplus \mathbf{V}_i^{\dagger}$. Here, $\mathbf{G}$ is the average of the $N$ final-layer features from the $N$-views' image backbones ($\mathcal{F}{\phi}$ in \cref{eq:keypoints_estimation}).
    \item \textbf{G\&J-emb} ($\mathbf{S^{+}}$). Similar to the MVP \cite{Wang2021MVP}, the $\mathbf{S^{+}}$ is initialized as the sum of global image feature $\mathbf{G}$, and vertex-level embedding vectors $\mathbf{S}^{J}$, as $\mathbf{S}_{i}^{+} = \mathbf{G} + \mathbf{S}_i^{J}$.
\end{enumerate}

We empirically find that the \textbf{J-emb} achieve the best performance 
(see \cref{para:v_posi_enc}\alphapara{G}).
Therefore, The feature of \V consists of two terms, namely (1) feature from projective aggregation, $\mathbf{F}$ (obtained via substituting \V for \ptinball in \cref{eq:projective_aggregation}) and  (2) $\mathbf{S}^{J}$ as the positional encoding of \V.
Following the convention used by \cite{Wang2021MVP,Hampali2022KeypointTS,lin2021metro}, we refer to this positional encoded feature as \textbf{query of} \V (denoted as \query).
Formally, we have $\mathbf{Q} = \mathbf{F} + \mathbf{S}^{J}$.

\subsubsection{Cross-set Point Transformer}
\vspace{-4 pt}
\label{sec:cross_set_point_transformer}
Given two pointclouds:
\begin{itemize}[leftmargin=10pt]
    \setlength\itemsep{-2 pt}
    \vspace{-4pt}
    \item $\mathbfcal{PC} = (\mathbf{\bar{P}}, \mathbf{F} )$, representing the frustum points' position \ptinball and  feature \pf aggregated from multi-view images;
    \item $\mathbfcal{VC} = (\mathbf{V}, \mathbf{Q} )$, representing the hand vertices' position \V and feature \query,
    \vspace{-4pt}
\end{itemize}
we want a set operator to progressively extract the inter-set relationship and update \Vcloud's prediction. 
The attention mechanism \cite{vaswani2017attention} is quite a natural and powerful choice for this task.
Inspired by the recent success of the self-attentive Point Transformer \cite{Zhao2021PointT}, 
we construct a cross-set Point Transformer to effectively capture relevant point features across different point clouds.  

Our Transformer only contains one decoder with multiple decoder layers in it. 
As shown in \cref{fig:architecture}\alphapara{(b)}, each decoder layer consists of three sequential modules: a self-attention to perform point-wise interaction in \Vcloud,
a vector cross-attention to perform point-to-set interaction from \Vcloud to \Pcloud,  
and a feed-forward network to regress a offset on \Vcloud's position (\V).
This offset will be used to refine the input \V from the previous layer.
When the Transformer reaches its final output stage, the \Vcloud's position will also reaches its convergence based on the learnable weights in the second stage.
The recursive form of each decoder layer is: $\mathbfcal{VC}_l = \mathcal{D}_{l} (\mathbfcal{PC}, \mathbfcal{VC}_{l-1})$, where  
$\mathcal{D}_l$ is the $l$-th decoder layer.
The point attention between  \Vcloud and \Pcloud is vector attention \cite{Zhao2020ExploringSF, Zhao2021PointT}, 
in which the attention will be applied in a local neighbor (\ie the $k$ nearest neighbor) of each query point (see \cref{fig:transformer_decoder_knn}).
We formulate the three modules in each $\mathcal{D}_l$ as:
\begin{equation}
    \setlength{\abovedisplayskip}{0pt}
    \setlength{\belowdisplayskip}{0pt}
    \small
    \begin{aligned}
    &\mathbf{Q}^{\star} =  \textit{self-attention}(\mathbf{Q}),\\
    &\widetilde{\mathbf{Q}}_i = \sum_{\mathbf{F}_j \in \mathbfcal{X}_i} \text{SM} \big( \gamma( \alpha (\mathbf{Q}^{\star}_i) - \beta(\mathbf{F}_j)) + \bm{\delta}) \big) \odot (\psi(\mathbf{F}_j) + \bm{\delta}),\\
    &\widetilde{\mathbf{V}}_i = \mathbf{{V}}_i + \text{FFN}(\widetilde{\mathbf{Q}}_i),  
    \label{eq:point_transformer} 
    \end{aligned}
\end{equation}
where $\mathbfcal{X}_i$ is a subset of \Pcloud, which collect points in the $k$ nearest neighbor of the $i$-th hand vertex: $\mathbf{V}_i$.
The $\bm{\delta}$ is the position encoding for point cloud, 
$\bm{\delta} = \theta(\mathbf{V}_i - \mathbf{\bar{P}}_j)$.
The $\alpha, \beta, \gamma, \psi, \theta$ are learnable functions (\eg MLP),  
$\odot$ is the Hadamard product, FFN is the feed-forward network, 
and SM is the \textit{softmax} operation.  
Through collecting all the $\widetilde{\mathbf{V}}_i$ and $\widetilde{\mathbf{Q}}_i$, we can get the output $\mathbfcal{VC}_l$ of the current decoder layer $\mathcal{D}_l$.
\del{In summary, \method's second stage $f_{\bm{\omega}}(\mathbf{X}, \mathbfcal{I})$ can be viewed as a set of learnable parameters and functions.  $f_{\bm{\omega}}:= \{ \bm{q}_{\textit{w}}, f_{m},  \mathcal{G},  \mathcal{D}_1, ..., \mathcal{D}_{L} \}$.}

\del{
\subsection{Learning Objectives}
Apart from the learning objectives $\mathcal{L}$ described in \cref{eq:overall_objectives},
we also adopt several auxiliary loss terms catering to mesh reconstruction, named as follows. 
The 2D projection loss $\mathcal{L}_{\mathbf{v}\text{2D}}$, the mesh normal loss $\mathcal{L}_{N}$, and edge loss $\mathcal{L}_{E}$. 
The $\mathcal{L}_{\mathbf{v}\text{2D}}$ penalize the pixel offset of \V's projection in $\mathit{l}$2 form. 
The $\mathcal{L}_{N}$ and $\mathcal{L}_{E}$ helps to smooth mesh reconstruction, as in \cite{Wang2018Pixel2Mesh}. 
The final loss term is thus defined as:
\begin{equation}
    \setlength{\abovedisplayskip}{5pt}
    \setlength{\belowdisplayskip}{5pt}
    \mathcal{L}_{\textit{\method}} =  \mathcal{L} + \mathcal{L}_{\mathbf{v}\text{2D}} +  \mathcal{L}_{E} + 0.1 \mathcal{L}_{N}.
    \label{eq:final_objectives} 
\end{equation}
}

\begin{table*}[t]
    \captionsetup{width=.95\linewidth}
    \renewcommand\arraystretch{1.0}
    \centering{
        \resizebox{0.97\linewidth}{!}
        {
            \setlength{\tabcolsep}{8pt}
        {
            \begin{tabular}{c|c|c|l|cccc|cccc}

            \shline
            \multicolumn{3}{c|}{} & \multirow{2}*{Methods} & \multicolumn{4}{c|}{Hand vertices} & \multicolumn{4}{c}{Hand keypoints} \\
            \multicolumn{3}{c|}{} &  & MPVPE$\downarrow$ & RR-V$\downarrow$ & PA-V$\downarrow$ & AUC-V$\uparrow$ & MPJPE$\downarrow$ & RR-J$\downarrow$ & PA-J$\downarrow$ & AUC-J$\uparrow$ \\
            \shline

            \multirow{7}*{\rotatebox{90}{DexYCB-MV}} & \texttt{1} & ours & \textbf{POEM} & \textbf{6.13}  & \textbf{7.21}  & \textbf{4.00}  & \textbf{0.70} &  \textbf{6.06} & \textbf{7.30} & \textbf{3.93}  & \textbf{0.68}   \\
            \hhline{~-----------}
            & \texttt{2} & \multirow{3}*{A} &  MVP \cite{Wang2021MVP} & 9.77 & 12.18 & 8.14 &0.53 & 6.23 & 9.47 & 4.26 & 0.69 \\
            & \texttt{3} &  & PE-Mesh-TR  & 7.41 & 8.67 & 4.70 & 0.64 & 7.49 & 8.87 & 4.76 & 0.64 \\
            & \texttt{4} &  & FTL-Mesh-TR & 8.75 & 9.80 & 5.75 & 0.59 & 8.66 & 9.81 & 5.51 & 0.59  \\
            \hhline{~~----------}
            & \texttt{8} & C & \method w/o pt. & 7.63 & 8.94 & 5.48 & 0.63 & 7.20 & 8.58 & 4.89 & 0.65 \\
            & \texttt{9} & D & \method w/o Proj. & 6.57 & 7.69 & 4.42 & 0.68 & 6.54 & 7.82 & 4.37 & 0.67 \\
            \hhline{~~----------}
            & \texttt{10} & F & Multi-view Fit. & 7.33 & 8.71 & 5.29 & 0.65 & 7.22 & 8.77 & 5.19 & 0.65 \\
            \midrule

            \multirow{7}*{\rotatebox{90}{\small{HO3D-MV}}} & \texttt{11} & ours & \textbf{POEM} &  \textbf{17.2} & \textbf{21.45} & \textbf{9.97}  & \textbf{0.66}  & \textbf{17.28}  & \textbf{21.94}  & \textbf{9.60}  & \textbf{0.63}  \\
            \hhline{~-----------}
            & \texttt{12} & \multirow{3}*{A} & MVP \cite{Wang2021MVP} & 20.95 & 27.08 & 10.04 & 0.59 & 18.72 & 24.90 & 10.44 & 0.60  \\
            & \texttt{13} &  & PE-Mesh-TR & 23.49 & 29.19 & 11.31 & 0.55 & 23.94 & 30.23 & 11.67 & 0.54 \\
            & \texttt{14} &  & FTL-Mesh-TR & 24.15 & 33.53 & 10.56 & 0.53 & 24.66 & 34.74 & 10.76 & 0.52 \\
            \hhline{~~----------}
            & \texttt{15} & C & \method w/o pt. & 19.26  & 24.32 & 12.45 & 0.62 & 18.20 & 23.80 & 10.56 & 0.63\\
            & \texttt{16} & D & \method w/o proj. & 18.83 & 22.26 & 10.83 & 0.63 & 18.48 & 22.73 & 10.39 & 0.63 \\
            & \texttt{17} & E & \method w/o ${\Delta \mathbf{v}}$ & 19.34 & 24.27 & 11.18 & 0.62 & 19.42 & 25.00 & 10.71 & 0.60 \\
            \midrule
            
            \multirow{8}*{\rotatebox{90}{\small{OakInk-MV}}} & \texttt{18} & ours & \textbf{POEM} ($\mathbf{S}^{J}$) & \textbf{6.20} & \textbf{7.63} & \textbf{4.21} & \textbf{0.70} & \textbf{6.01} & \textbf{7.46} & \textbf{4.00} & \textbf{0.69}\\
            \hhline{~-----------}
            & \texttt{19} & \multirow{3}*{A} & MVP \cite{Wang2021MVP} & 9.69 & 11.75 & 7.74 & 0.53 & 7.32 & 9.99 & 4.97 & 0.64 \\
            & \texttt{20} &  & PE-Mesh-TR & 8.34 & 9.67 & 5.75 & 0.60 & 8.18 & 9.59 & 5.42 & 0.61 \\
            & \texttt{21} &  & FTL-Mesh-TR & 9.28 & 10.88 & 6.61 & 0.56 & 8.89 & 10.66 & 6.01 & 0.58 \\
            \hhline{~~----------}
            & \texttt{22} & D & \method w/o proj & 6.42 & 7.82 & 4.50 & 0.69 & 6.25 & 7.84 & 4.28 &  0.68 \\
            & \texttt{23} & E & \method w/o ${\Delta \mathbf{v}}$ & 6.56 & 8.04 & 4.63 & 0.69 & 6.32 & 7.99 & 4.32 & 0.67 \\
            \hhline{~~----------}
            & \texttt{24} & \multirow{2}*{G} & $\mathbf{S^{+}}$ as pos-enc. & 6.23 & 7.63 & 4.30 &0.70 & 6.05 & 7.65 & 4.09 &0.69 \\
            & \texttt{25} &  & $\mathbf{S}^{G}$ as pos-enc. & 6.25 & 7.65 & 4.33 & 0.70& 6.05 & 7.66 & 4.10 & 0.69\\
            \shline

    \end{tabular}}}}
    \vspace{-5 pt}
    \caption{
        Quantitative results (mm) of evaluations \textbf{A} to \textbf{F}. The AUC are computed on the evaluation metrics of MPVPE and MPJPE. The thresholds of AUC vary from datasets. \ie 0-20 $mm$ for DexYCB-MV and OakInk-MV, and 0-50 $mm$ for HO3D-MV.
    }\vspace{-8 pt}
    \label{tab:quantitative}
    
\end{table*}

\section{Experiments and Results}
\subsection{Datasets}
\vspace{-0.6em}
\qheading{DexYCB.}
It contains 582K images of hand grasping objects \cite{Chao2021DexYCB}. These images consist of the observations from 8 cameras. 
To construct a split mode for multi-view task, we follow its official `S0' split on train/val/test sets and filter out the frames on left hand. For each frame ID, we collect 8 images from total 8 cameras. These 8 images, along with their annotations, consist of one multi-view (\mv) frame. We name the DexYCB with our multi-view split as \dexycb. In all, \dexycb contains 25,387 \mv frames (that is, 203,096 monocular frames) in training set, 1,412 \mv frames in validation and 4,951 in testing set. 

\qheading{HO3D.}
HO3D (version 3) \cite{hampali2021ho3dv3} contains 103,462 images capturing hand-object interaction from up-to 5 cameras. 
Frames of same sequence but different cameras may be scattered in different split sets.  
To enable HO3D supporting multi-view task, we only select sequences with full 5 camera observations, and construct a HO3D-MV upon them. 
There are total 7 sequences in HO3D satisfy multi-view requirements. We select sequences with serial: `ABF1',`BB1', `GSF1', `MDF1' and `SiBF1' as training set, and the remain `GPMF1' and `SB1' as testing set. 
In total, there are 9,087 \mv frames (that is 45,435 monocular frames) in training set and 2,706 in testing set. 

\qheading{OakInk.}
OakInk-Image \cite{Yang2022OakInk} is a dataset of hand manipulating objects. It contains 230K images from 
the observation of 4 cameras. We follow the official `SP2' (objects split) and construct 
the OakInk-MV. Each frame in OakInk-MV contains the images from 4 cameras. 
Nearly a quarter of sequences in OakInk contain two person handing over an object (with two hands captured in  image). For these sequences, we train and test each hand separately.  
OakInk-MV has 58,692 \mv frames (that is, 234,768 monocular frames) in training set and 19,909 \mv frames in testing set.

\subsection{Evaluation}\label{sec:evaluation}
We report the MPJPE and MPVPE ($mm$), standing for the mean per keypoint (joint) and per vertex position error, respectively. Notably, since the MANO provides a pre-trained mapping from \V to \X, 
The reported keypoints \X is a by-product of the final vertex \V from the second stage.
Beside, we also inspect the MPVPE and MPVPE in a root-relative (RR) system and under the Procrustes analysis (PA).
Additionally, we evaluate the percentage of correct keypoints under a range of threshold by measuring the area under the percentage curve (AUC), which tells us the model's discriminative ability on localizing keypoints.

\vspace{-0.2em}\qparagraph{A. \method\xspace-vs- SOTA.}
\method targets on single hand reconstruction under the multi-view RGB observations. We find that the MVP\cite{Wang2021MVP}, which directly regress mesh parameters catered to a skinning body model (SMPL\cite{Loper2015SMPL} and MANO\cite{romero2017embodied}) using vision Transformer, is highly relevant to our model. MVP also employs another module to associate keypoints among different persons. In our experiments, we only compare \method with MVP \wrt the single-body reconstruction. 
Apart from MVP, we also simulate several SOTA methods for multi-view hand mesh reconstruction. 
The purpose of the simulation is to combine the SOTA architecture in monocular hand reconstruction with the most advanced fusion algorithms adopted by multi-view settings.
With this in mind, we first explore the advanced model in the field of 3D detection in autonomous driving, as operating neural network on multi-view setting is well-explored in this field.
The recent method: PETR \cite{Liu2022PETR} 
encoded the 3D position embedding (PE) of camera frustum into 2D image features, 
fused those position-embedded features with learnable objects' queries (corresponding to the \query in our model), and finally transferred those objects' queries to the objects' 3D position via a DETR \cite{Carion2020DETR} decoder.
In terms of monocular hand mesh reconstruction, the Mesh Transformer \cite{lin2021metro} is a representative method that adopts an image-based self-attentive Transformer to directly regress all vertices on the hand mesh.
Therefore, for simulating multi-view hand mesh reconstruction, we utilize the feature-position encoder of PETR to process multi-view image features and a DETR-like vision Transformer to regress the 3D vertices on hand mesh (based on the design in \cite{lin2021metro}). We denote this method ``PE-Mesh-TR''.
Similarly, we simulate the ``FTL-Mesh-TR'', which utilizes the feature transform layer (FTL) \cite{Remelli2020LightweightM3} to fuse multi-view features and a mesh-adapted DETR decoder for mesh regression. Our \method outperforms these methods in all metrics.

\qparagraph{B. \method's Two-stage Design.}\label{para:poem_two_stage}
Recalling the problem formulation in \cref{sec:formulation}-\cref{eq:formulation}, the method MVP, PE-Mesh-TR and FTL-Mesh-TR can be represented as $P(\mathbf{X}, \mathbf{V} | \mathbfcal{I})$, where the hand vertices and keypoints are directly regressed from their models at the output stage. In contrast, our method is explicit formulated as two-stage model. Their comparisons are shown in row \linenb{1} \vs \linenb{2-4}, \linenb{11} \vs \linenb{12-14} and \linenb{18} \vs \linenb{19-21}.

\qparagraph{C. \method without Embedded Point.}
We also simulate a two-stage version of \method but \textbf{without} embedded points design, named 
``POEM w/o pt''. We sequentially concatenate the first stage in \method (for predicting \X), the feature-position encoder, and Transformer decoder as in the  PE-Mesh-TR.
This lead inferior results, shown in row \linenb{1} \vs \linenb{8} and \linenb{11} \vs \linenb{15}.

\qparagraph{D. \method without Projective Aggregation.}
We investigate the performance of two feature-to-points embedding methods in \cref{sec:feats_to_points_assignment}. The results of the position embedded aggregation are denoted as 
``\method w/o proj''. Comparison between row \linenb{1} \vs \linenb{9}, \linenb{11} \vs \linenb{16} and \linenb{18} \vs \linenb{22} show that the later projective aggregation is effective.

\qparagraph{E. \method without Progressive \Vcloud Update.}
We remove all the FFN in the decoder layer only except the last one. 
The final \V is thus a single-step prediction (``\method w/o ${\Delta \mathbf{V}}$''). The comparison can be found in row \linenb{11} \vs \linenb{17} and \linenb{18} \vs \linenb{23}, showing that removing 
the progressive \Vcloud's update leads to inferior results.

\qparagraph{F. \method\xspace-vs- Multi-view Mesh Fitting.}\label{para:mesh_from_fit}
Multi-view pose estimation is a common practice for datasets to obtain hand's ground-truth \cite{hampali2020ho3dv2, Qin2022DexMV}. 
For instance, HOnnotate \cite{hampali2020ho3dv2} aggregated visual cues from several upstream vision tasks to acquire the hand model automatically. In a similar way, we also fit a 3D hand mesh (MANO) to the multi-view 2D predictions.
Our fitting objectives are three-fold: 
\begin{enumerate*}[label={\arabic*)}]
    \item 2D hand keypoints;
    \item silhouette;
    \item hand's anatomical constraints \cite{yang2021cpf};
\end{enumerate*}
The pseudo ground-truth of 2D keypoints and silhouette are obtained from a pretrained model \cite{chen2021camera}. 
We report the final results in row \linenb{1} \vs \linenb{10}. 
\method outperforms the time-consuming fitting-based methods for auto-labeling.

\qparagraph{G. \method: \V's Positional Encoding.}\label{para:v_posi_enc}
We explored three different strategies for constructing the position encoding (pos-enc) of \V, namely $\mathbf{S}^{J}$, $\mathbf{S}^{G}$, and $\mathbf{S}^{+}$, as described in \cref{para:feats_to_v_assignment}\alphapara{B}. The results in row \linenb{18} \vs \linenb{24,25} show that the joint/vertex-level embedding ($\mathbf{S}^{J}$ ) best fit for our model.

\qparagraph{H. Decoders, Cameras, and \textit{k} Neighbors.}
We further examine the performance of \method on varying the number of decoder, numbers of cameras, and numbers of $k$ nearest neighbors, and qualitatively assess the \method's reconstruction results. Please refer to the \supp for more details.

\section{Discussion}
\vspace{-1mm}
\qheading{Limitation.} First, \method is not able to distinguish keypoint from different hands.
Hence, it only supports single hand reconstruction. 
Second, to fully exploit its power, the projective aggregation expects those cameras to have an apparent spatial difference. Otherwise, it will collapse to the aggregation of 2D features with adjacent coordinates.

\qheading{Conclusion.}
In this paper, we propose \method, which addresses the multi-view hand reconstruction using point representation. \method directly operates on points through two novel designs, that are point-based feature fusion and cross-set point attention. 
Experiments on three datasets show the effectiveness of such designs. 
Though targeting on hand's reconstruction, the virtue brought from \method also opens a path toward the general multi-view-based object reconstruction, which we leave as a future work.

\noindent\rule[0.3ex]{\linewidth}{1.0pt}

\qheading{Acknowledgments.}  
This work was supported by the National Key R\&D Program of China (No. 2021ZD0110704), Shanghai Municipal Science and Technology Major Project (2021SHZDZX0102), Shanghai Qi Zhi Institute, and Shanghai Science and Technology Commission (21511101200).
We thank Anran Xu, Jiefeng Li and Kailin Li for their help, discussion and feedback.

\begin{figure*}[ht]
  \begin{center}
    \includegraphics[width=1.0\linewidth]{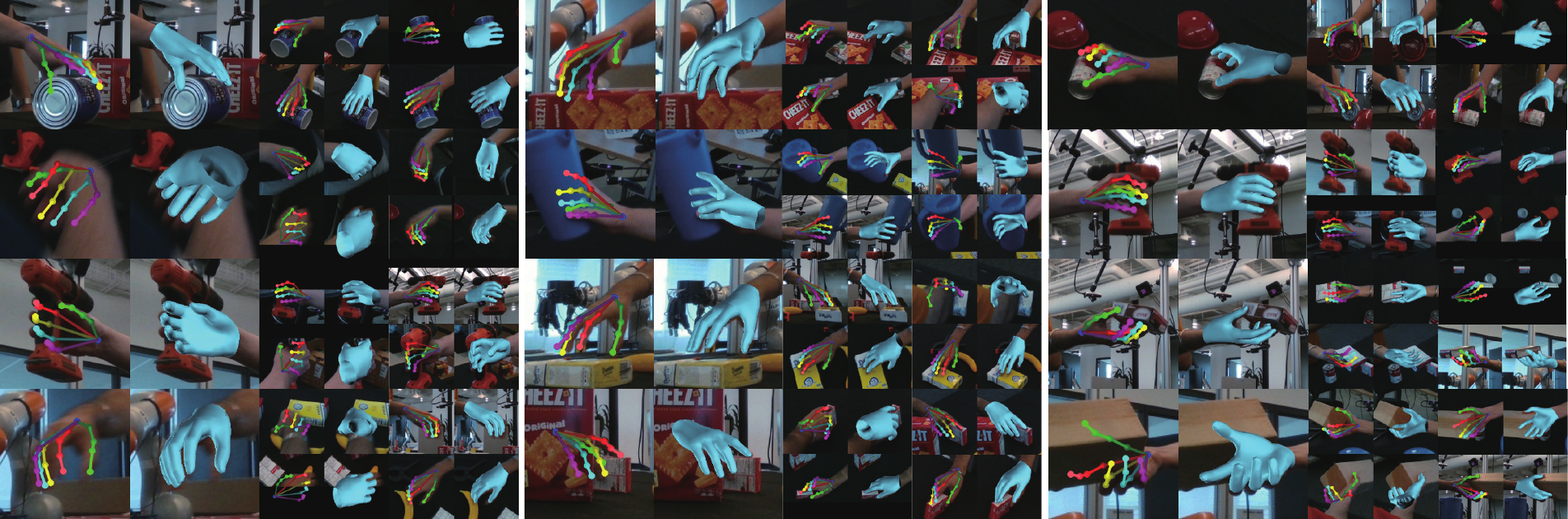}\vspace{-0.5em}
    \caption{Qualitative results on DexYCB-MV testing set.} \label{fig:dexycb_viz}
  \end{center}

  \begin{center}
    \includegraphics[width=1.0\linewidth]{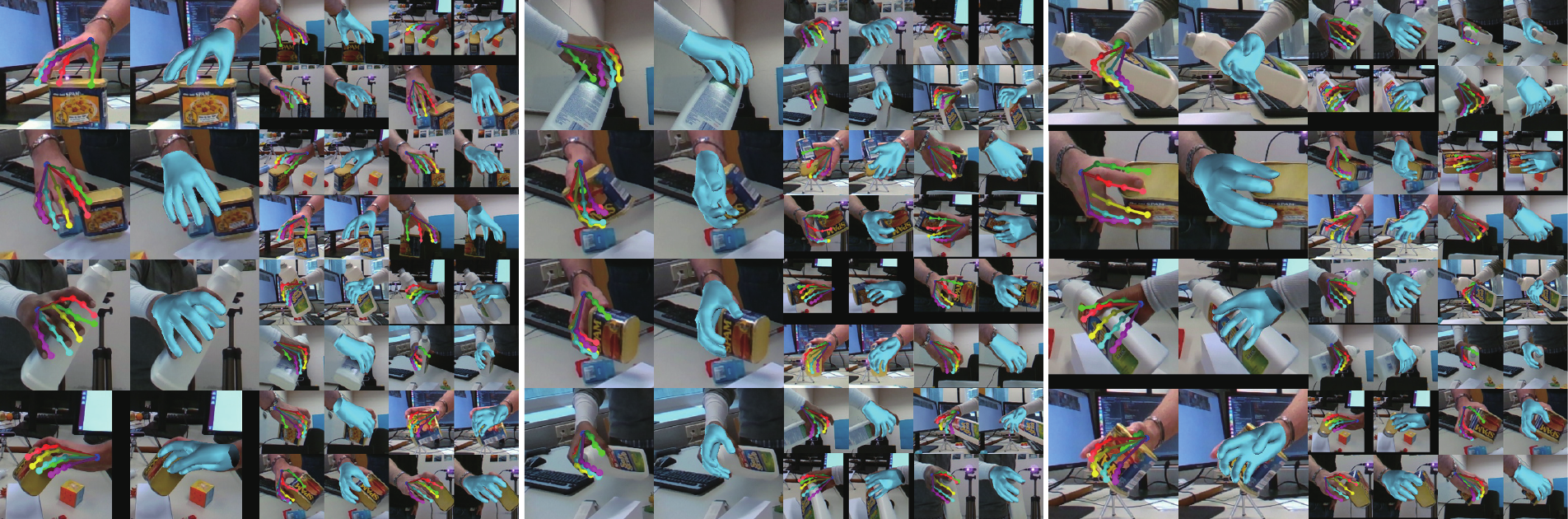}\vspace{-0.5em}
    \caption{Qualitative results on HO3D-MV testing set.} \label{fig:ho3d_viz}
  \end{center}

  \begin{center}
    \includegraphics[width=1.0\linewidth]{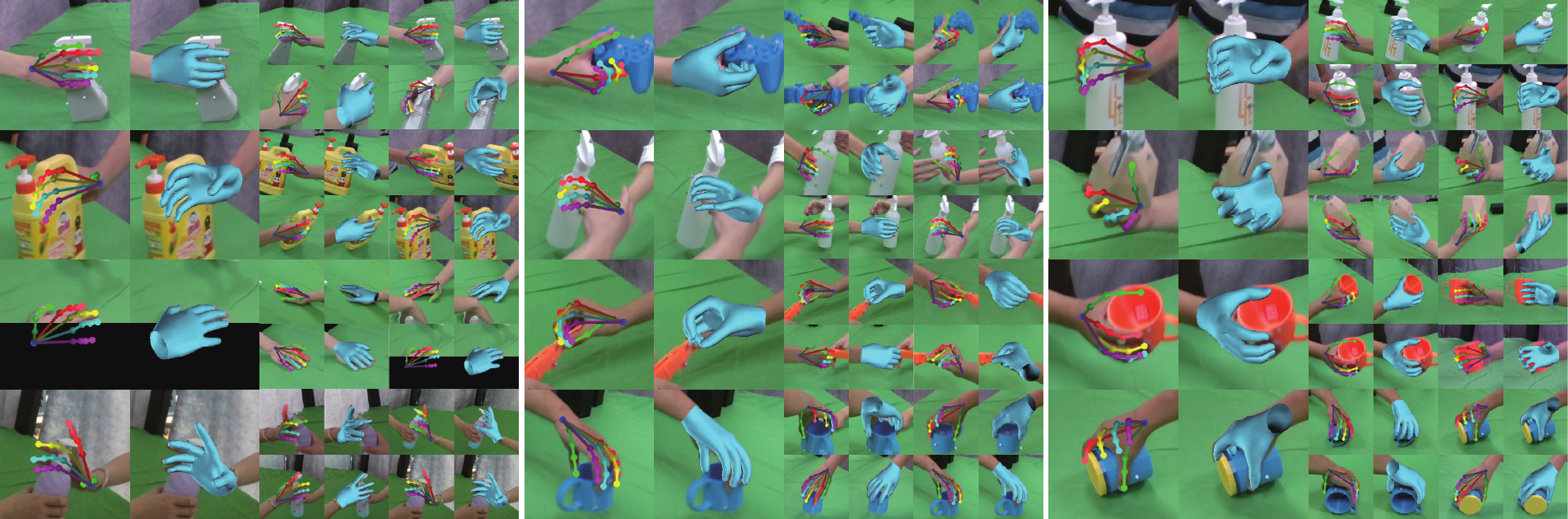}\vspace{-0.5em}
    \caption{Qualitative results on OakInk-MV testing set.} \label{fig:oakink_viz}
  \end{center}
  \vspace{-1.0em}
\end{figure*}

\clearpage
{
\balance
\begin{appendices}\label{appendices}
\section{Implementation Details}\label{sec:supp_impl_details}

All the experiments are developed using PyTorch library and  are conducted on a machine with 4 NVIDIA A10 GPUs (24GB RAM).
These experiments use the batch size of 16 and total 100 epochs of training. 
The learning rate is set to $1 \times 10^{-4}$ and decayed by a factor of 0.1 at the 70 epochs. 
All the experiments related to the multi-view settings use a same CNN backbone: ResNet34 \cite{resnet}. 
The vision Transformer is initialized with \textit{xavier} uniform distribution and the CNN backbone is initialized with ImageNet \cite{imagenet} pre-trained weights.

Regarding to the experiments in \method, the radius of ball query is set to be 0.2$m$ around the center of hand and total $S = 2048$ points (number of points in \ptinball) are sampled within this range. The number $k$ of nearest neighbors is set to be 16 in the cross-set point Transformer.  

We apply standard image augmentation techniques to train our model, including random center offset, scaling, and color jittering. Additionally, we apply random rotation. However, rotational augmentation in the multi-view setting differs from that in the single-view setting. In single-view training, a rotation on the image corresponds to the same rotation on the 3D hand model. However, in multi-view training, a rotation on the image is represented as left-multiplication of the rotation on the camera extrinsic matrix.

\section{More evaluations}\label{sec:supp_evaluation}

\cqheading{Number of Decoder Layers.} We examine the performance of \method on varying the numbers of decoder layers in its point Transformer. 
The results on HO3D-MV are shown in \cref{tab:exp_abl_appx} rows \texttt{\small 1-4}, where the ``d1'' indicate only use one decoder layer. We find that using 6 decoder layers achieves the best performance.

\cqheading{Numver of Camera Views.} We evaluate the performance of \method on varying the numbers of cameras in the multi-view setting.
The results on HO3D-MV are shown in \cref{tab:exp_abl_appx} rows \texttt{\small 5-8}, where the ``c2'' indicates only use two cameras. The results show \method can effectively fuse the features from different camera frustums and thus boost the performance when the number of cameras increases.

\cqheading{Number of \textit{k} Nearest Neighbors.} 
We evaluate the performance of \method on varying  the number of $k$ from 4, 8, 16, to 32.
A value of $k=32$ with a batch size of 16 almost exhausts the memory of the A10 GPU.
\cref{tab:exp_abl_appx} rows \texttt{\small 9-12} show that larger $k$ could lead to better results. But the computation cost also increases with it.
We use $k$=16 for the trade-off between the cost and performance.
\begin{table}[H]
    \tiny
    \centering{
        \resizebox{1.0\linewidth}{!}
        {
            \setlength{\tabcolsep}{3pt}
        {
            \begin{tabular}{c|c|ccc|ccc}
            \shline
            & \multirow{2}*{\textbf{Exp}} & \multicolumn{3}{c|}{Joints} & \multicolumn{3}{c}{Vertices} \\
            & & MPJPE & RR-J & PA-J & MPVPE & RR-V & PA-V   \\
            \hline
             \multirow{8}*{\rotatebox{90}{HO3D}}& d1 &  19.29 & 24.49 & 11.19   & 19.17  & 23.86  & 11.99  \\
             & d2 &  19.00 & 24.34 & 11.22   & 18.72  & 23.65  & 11.81  \\
             & d4 &  18.81 & 24.33 & 10.79 & 18.52 & 23.58 & 11.27   \\
             & d6 &  \textbf{17.55} & \textbf{22.59} & \textbf{9.83}   & \textbf{17.56}  & \textbf{23.07}  & \textbf{9.40}  \\
            \hhline{~-------}
             & c2 &  28.82 & 38.53 & 20.85   & 28.86  & 38.27  & 22.43  \\
             & c3 &  26.29 & 40.55 & 17.78   & 27.72  & 39.78  & 21.65  \\
             & c4 &  20.82 & 26.77 & 12.18   & 20.55  & 25.84  & 12.69  \\
             & c5 &  \textbf{19.22} & \textbf{24.97} & \textbf{11.17}    & \textbf{18.93}  & \textbf{24.22}  & \textbf{11.59}   \\
            \hline
             \multirow{4}*{\rotatebox{90}{OakInk}}& k4 & 6.34 & 8.08 & 4.40 & 8.33 & 9.75 & 6.87 \\
            & k8 & 6.39 & 8.07 & 4.42 & 8.16 & 9.57 & 6.64 \\
            & k16  & \textbf{6.34} & \textbf{8.02} & \textbf{4.37} & 8.08 & 9.53 & 6.57\\
            & k32 & 6.36 & 8.02 & 4.39 & \textbf{7.93} & \textbf{9.39} & \textbf{6.36} \\
            \shline
        \end{tabular}}}}
        \vspace{-1em}
        \caption{Performance of \method on varying the number of decoder layers, cameras, and \textit{k} nearest neighbors in the multi-view setting. The best results are highlighted in \textbf{bold}.}
        \label{tab:exp_abl_appx}
\end{table}

\vspace{-0.5em}
\cqheading{Inference Time.} 
\cref{tab:exp_test_appx} compares the inference time and model parameters of four methods: \method, MVP, PE-Mesh-TR, and the multi-view mesh fitting. The inference time is calculated as an average feed-forward time of one multi-view sample on one GPU.
\begin{table}[h]
\centering
\resizebox{0.8\linewidth}{!}{
    \setlength{\tabcolsep}{5pt}{
    \begin{tabular}{l|cccc}
    \toprule
        Model &  POEM & MVP & PE-Mesh-TR & Fit.\\ 
    \midrule
        time (s)  & 0.067  & 0.055 & 0.035 & 9.89 \\
        params(M) &  117 & 144 & 124  &  -- \\ 
    \bottomrule
    \end{tabular}
    }}\vspace{-2mm}
    \caption{Inference time and model parameters.}\vspace{-2mm}
    \label{tab:exp_test_appx}
\end{table}

\vspace{-1em}\section{Qualitative results}\label{sec:supp_qualitative}
We demonstrate more qualitative results of \method on the three datasets in \cref{fig:dexycb_viz,fig:ho3d_viz,fig:oakink_viz}.  From top to bottom we plot the results on DexYCB-MV, HO3D-MV and OakInk-MV dataset.
For each multi-view frame, we draw its result from 5 different views. One of the views is in normal size and the other four views are half size.

\end{appendices}
}

\clearpage
{\small
    \balance
    \bibliographystyle{configs/ieee_fullname}
    \bibliography{egbib}
}

\end{document}